\documentclass[hidelinks,10pt,twocolumn,letterpaper]{article}
\usepackage{cvpr}
\usepackage{times}
\usepackage{epsfig}
\usepackage{graphicx}
\usepackage{amsmath}
\usepackage{amssymb}
\usepackage{amssymb}
\usepackage{microtype}
\usepackage{booktabs}
\usepackage{tabularx}
\usepackage{subfig}
\usepackage{floatrow}

\newfloatcommand{capbtabbox}{table}[][\FBwidth]

\usepackage[breaklinks=true,bookmarks=false,colorlinks=true]{hyperref}

\begin{document}

\title{Disentangling Motion, Foreground and Background Features in Videos}

\author{\small Xunyu Lin\thanks{This work was developed during an exchange period of Xunyu Lin at the Universitat Politecnica de Catalunya.}
\\
{\small Beihang University}\\
{\small Beijing, China}\\
{\tt\small xunyulin2017@outlook.com}
\and
{\small V{\'i}ctor Campos}\\
{\small Barcelona Supercomputing Center}\\
{\small Barcelona, Catalonia/Spain}\\
{\tt\small victor.campos@bsc.es}
\and
{\small Xavier Giro-i-Nieto}\\
{\small Universitat Politecnica de Catalunya}\\
{\small Barcelona, Catalonia/Spain}\\
{\tt\small xavier.giro@upc.edu}
\and
{\small Jordi Torres}\\
{\small Barcelona Supercomputing Center}\\
{\small Barcelona, Catalonia/Spain}\\
{\tt\small jordi.torres@bsc.es}
\and
{\small Cristian Canton Ferrer}\\
{\small Microsoft}\\
{\small Redmond (WA), USA}\\
{\tt\small cristian.canton@microsoft.com}
}


\maketitle

\begin{abstract}
This paper introduces an unsupervised framework to extract semantically rich features for video representation. 
Inspired by how the human visual system groups objects based on motion cues, we propose a deep convolutional neural network that disentangles motion, foreground and background information.
The proposed architecture consists of a 3D convolutional feature encoder for blocks of 16 frames, which is trained for reconstruction tasks over the first and last frames of the sequence.
A preliminary supervised experiment was conducted to verify the feasibility of proposed method by training the model with a fraction of videos from the UCF-101 dataset taking as ground truth the bounding boxes around the activity regions.
Qualitative results indicate that the network can successfully segment foreground and background in videos as well as update the foreground appearance based on disentangled motion features.
The benefits of these learned features are shown in a discriminative classification task, where initializing the network with the proposed pretraining method outperforms both random initialization and autoencoder pretraining. 
Our model and source code are publicly available at \url{https://imatge-upc.github.io/unsupervised-2017-cvprw/} .

\end{abstract}
\section{Introduction}


\par Unsupervised learning has long been an intriguing field in artificial intelligence. Human and animal learning is largely unsupervised: we discover the structure of the world mostly by observing it, not by being told the name of every object, which would correspond to supervised learning \cite{lecun15nature}. 
A system capable of predicting what is going to happen by just watching large collections of unlabeled video data needs to build an internal representation of the world and its dynamics \cite{mathieu2015deep}. When considering the vast amount of unlabeled data generated every day, unsupervised learning becomes one of the key challenges to solve in the road towards general artificial intelligence.
%
%

Based on how a human would provide a high level summary of a video, we hypothesize that there are three key components to understand such content: namely \textit{foreground}, \textit{motion} and \textit{background}. These three elements would tell us, respectively, what the main objects in the video are, what they are doing and where their location. We propose a framework that explicitly disentangles these three components in order to build strong features for action recognition, where the supervision signals can be generated without requiring from expensive and time consuming human annotations.
The proposal is inspired by how infants who have no prior visual knowledge tend to group things that move as connected wholes and also move separately from one another \cite{elizabeth90cognitivemotion}. Based on this intuition, we can build a similar unsupervised pipeline to segment foreground and background with global motion, i.e. the rough moving directions of objects. Such segmented foregrounds across the video can be used to model both the global motion (e.g.~transition or stretch) and local motion (i.e.~transformation of detailed appearance) from a pair of foregrounds at different time steps. 
Since background motion is mostly given by camera movements, we restrict the use of motion to the foreground and rely on appearance to model the former.

The contributions of this work are two-fold: (1) disentangling motion, foreground and background features in videos by human alike motion aware mechanism and (2) learning strong video features that improve the performance of action recognition task. 



\section{Related Work}




%
%



Leveraging large collections of unlabeled videos has proven beneficial for unsupervised training of image models thanks to the implicit properties they exhibit in the temporal domain, e.g. visual similarity between patches in consecutive frames \cite{wang2015unsupervised} and temporal coherence and order \cite{misra2016shuffle}. Since learning to predict future frames forces the model to construct an internal representation of the world dynamics, several works have addressed such task by predicting global features of future frames with Recurrent Neural Networks (RNN) \cite{srivastava2015unsupervised} or pixel level predictions by means of multi-scale Convolutional Neural Networks (CNN) trained with an adversarial loss \cite{mathieu2015deep}. The key role played by motion has been exploited for future frame prediction tasks by explicitly decomposing content and motion \cite{villegas2017decomposing} and for unsupervised training of video-level models \cite{luo2017motionprediction}. Similarly in spirit, separate foreground and background streams have been found to increase the quality of generative video models \cite{vondrick2016generating}.

Techniques exploiting explicit foreground and background segmentations in video generally require from expensive annotation methods, limiting their application to labeled data. However, the findings by Pathak et al. \cite{pathak2017learning} show how models trained on noisy annotations learn to generalize and perform well when finetuned for other tasks. Such noisy annotations can be generated by unsupervised methods, thus alleviating the cost of annotating data for the target task. In this work we study our proposed method by using manual annotations, whereas evaluating the performance drop when replacing such annotations with segmentations generated in an unsupervised manner remains as future work.

\section{Methodology}

\begin{figure*}[ht]
\centering
\includegraphics[width=\linewidth]{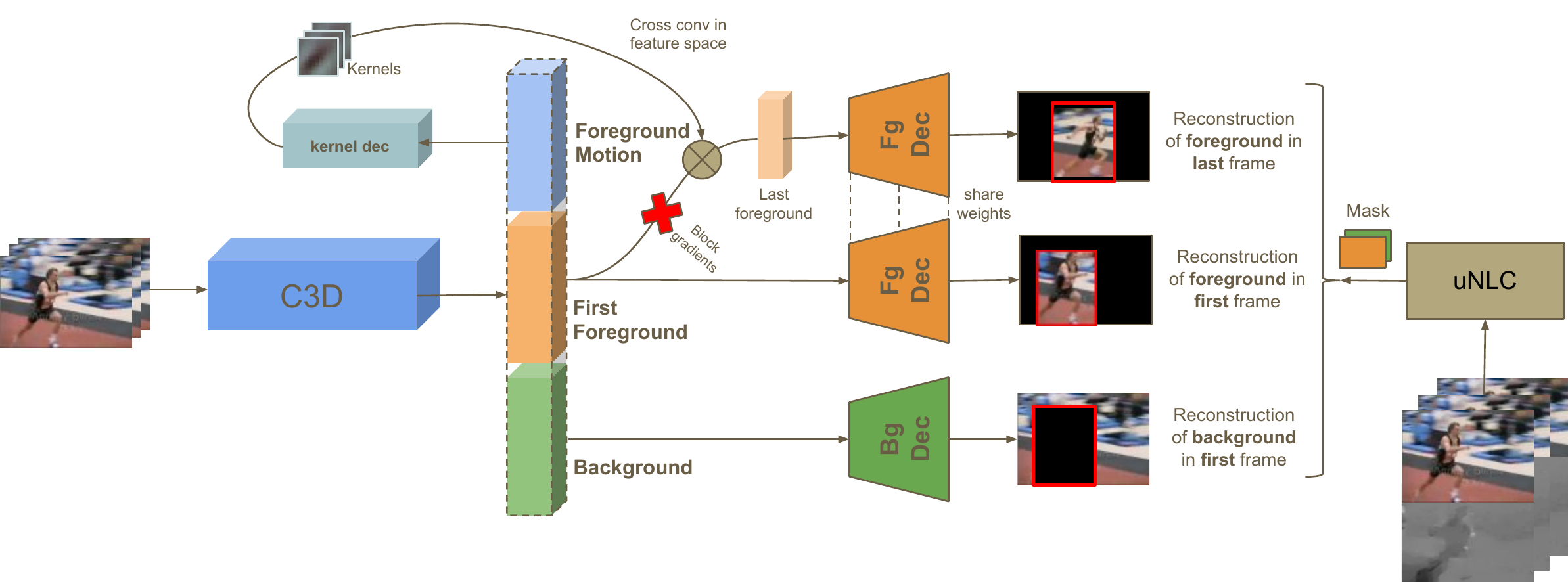}\\
\caption{System architecture. Please note that in this work, the masks used to generate ground truth are from manual annotations while uNLC will be utilized in our future work.}
\label{fig:architecture}
\end{figure*}


We adopt an autoencoder-styled architecture to learn features in an unsupervised manner.  The encoder maps input clips 
to feature tensors 
by applying a series of 3D convolutions and max-pooling operations \cite{du2015c3d}. Unlike traditional autoencoder architectures, the bottleneck features are partitioned into three splits which are then used as input for three different reconstruction tasks, as depicted in Figure \ref{fig:architecture}.

\textbf{Disentangling of foreground and background:} depending on the nature of the training data, reconstruction of frames may become dominated either by the foreground or background.  We explicitly split the reconstruction task to guarantee that none of the parts dominates over the other. Partitioned foreground and background features will be passed into two different decoders for reconstruction. While segmentation masks are often obtained by manual labeling, it is worth noting they can be obtained without supervision as well, e.g. by using methods based on motion perceptual grouping such as uNLC \cite{pathak2017learning}. The latter approach has proven beneficial for unsupervised pre-training of CNNs \cite{pathak2017learning}.

\textbf{Disentangling of foreground motion:} leveraging motion information can provide a boost in action recognition performance when paired with appearance models \cite{simonyan2014two}. We encourage the model to learn motion-related representations by solving a predictive learning task where the foreground in the last frame needs to be reconstructed from the foreground in the first frame. Given a pair of foregrounds at timesteps $t_1$ and $t_2$, namely $\left(f_{t_{1}}, f_{t_{2}} \right)$, we aim to estimate a function $M$ from motion features $m_{t_1\rightarrow t_2}$ throughout $t_1$ and $t_2$ that maps $f_{t_{_1}}$ to $f_{t_{2}}$ in deep feature space $G$: 

\begin{equation}
G \left( f_{t_{2}} \right) = M \left( G(f_{t_{1}}), m_{t_1 \rightarrow t_2} \right)
\end{equation}

Throughout this work, the space of encoded features is used for $G$, and $M$ is parametrized by a deterministic version of cross convolution \cite{visualdynamics16}. The foreground decoder weights are shared among all foreground reconstruction task. Gradients coming from the reconstruction of $f_{t_{2}}$ are blocked from backpropagating through $G( f_{t_{1}})$ during training to prevent $G( f_{t_{1}})$ from storing information about $f_{t_{2}}$. 


\textbf{Frame selection:} assuming that the background semantics stay close throughout the short clips, only the background in the first frame is reconstructed. First and last frames are chosen to perform foreground reconstruction, since they represent the most challenging pair in the clip.

\textbf{Loss function:} the model is optimized to minimize the L1 loss between the original frames and their reconstruction. In particular, the loss function is defined from a decomposition of the input video volume $x$ of $T$ frames into the foreground $x_{fg}$ and background $x_{bg}$ volumes:

\begin{equation}
\begin{split}
x_{fg} = x \cdot b_{fg} \\
x_{bg} = x \cdot (1- b_{fg})
\end{split}
\end{equation}

where $b_{fg}$ corresponds to a volume of binary values, so that $1$ correspond to foreground pixels and $0$ to the background ones.

This decomposition allows defining the reconstruction loss $L_{rec}(x)$ over the video volume $x$ as the sum of three terms:

\begin{equation}
\label{eq:loss}
L_{rec}(x) = L^{1}_{fg}(x) + L^{1}_{bg}(x) + L^{T}_{fg}(x)
\end{equation}

where the components $L^{1}_{fg}$, $L^{1}_{bg}$ and $L^{T}_{fg}$ represent the reconstruction loss for the first foreground and first background , and last foreground , respectively. These three terms are particularizations at the first ($t=1$) and last ($t=T$) frames of the generic foreground $L^{t}_{fg}(x)$ and background $L^{t}_{bg}(x)$ reconstructions losses:

\begin{equation}
L^{t}_{fg}(x) = \frac{1}{A^t}\sum_{i,j}{W^t[i,j] \cdot \left|\hat{x}^{t}_{fg}[i,j] - x_{fg}[i,j,t]\right|}
\end{equation}
\begin{equation}
L^{t}_{bg}(x) = \frac{1}{A^t}\sum_{i,j}{\left|\hat{x}^t_{bg}[i,j] - x_{bg}[i,j,t]\right|}
\end{equation}

where $\hat{x}^t$ denotes a reconstructed foreground/ background at time $t$, $A^t$ is the area of the reconstructed frame at time $t$, and $W^t$ is an element-wise weighting mask at time $t$ designed to leverage the focus between the foreground and background pixels:

\begin{equation}
W^{t}[i,j]= 
\begin{cases}
1 & \text{if } (i,j) \in \text{background}\\
\max\left[ 1, \frac{A^t_{bg}}{A^t_{fg}} \right] & \text{if } (i,j) \in \text{foreground}
\end{cases}
\end{equation}

During preliminary experiments, we observed that the reconstruction of the first foreground always outperformed the reconstruction of the last one by a large margin, given the increased difficulty of the latter task. In order to get finer reconstruction of the last foreground, we introduce an L2 loss $L_{feat}$ on $G(f{t_{2}})$. The pseudo ground truth for this task is obtained by getting first foreground features from the encoder fed with the temporally reversed clip. The final loss to optimize is the following:

\begin{equation}
L_{total}(x) = L_{rec}(x) + L_{feat}(x) 
\end{equation}

\section{Experimental setup}
Please note again we are showing results trained with ground truth masks to check the feasibility of our proposal and the pure unsupervised framework generating masks from uNLC \cite{pathak2017learning} remains as future work.

\textbf{Dataset:} there are 24 classes out of 101 in UCF-101 with localization annotations \cite{UCF101,THUMOS15}. Following \cite{pathak2017learning}, we first evaluate the proposed framework with supervised annotations and use the bounding boxes in the subset of UCF-101 for such purpose. Evaluating the proposal in weak annotations collected by means of unsupervised methods remains as future work. We follow the original splits of training and test set and also split 10\% videos out of the training set as validation set in order to perform early stopping and prevent the network from overfitting the training data.

\textbf{Training details:} videos are split into clips of 16 frames each. These clips are then resized to $128\times128$ and their pixel values are scaled and shift to $[ -1, 1]$. The clips are randomly temporally or horizontally flipped for data augmentation. Weight decay with rate of $10^{-3}$ is added as regularization. The network is trained for 125 epochs with Adam optimizer and a learning rate of $10^{-4}$ on batches of 40 clips.

\section{Results}

\begin{figure*}[ht]
\centering
\includegraphics[width=\linewidth]{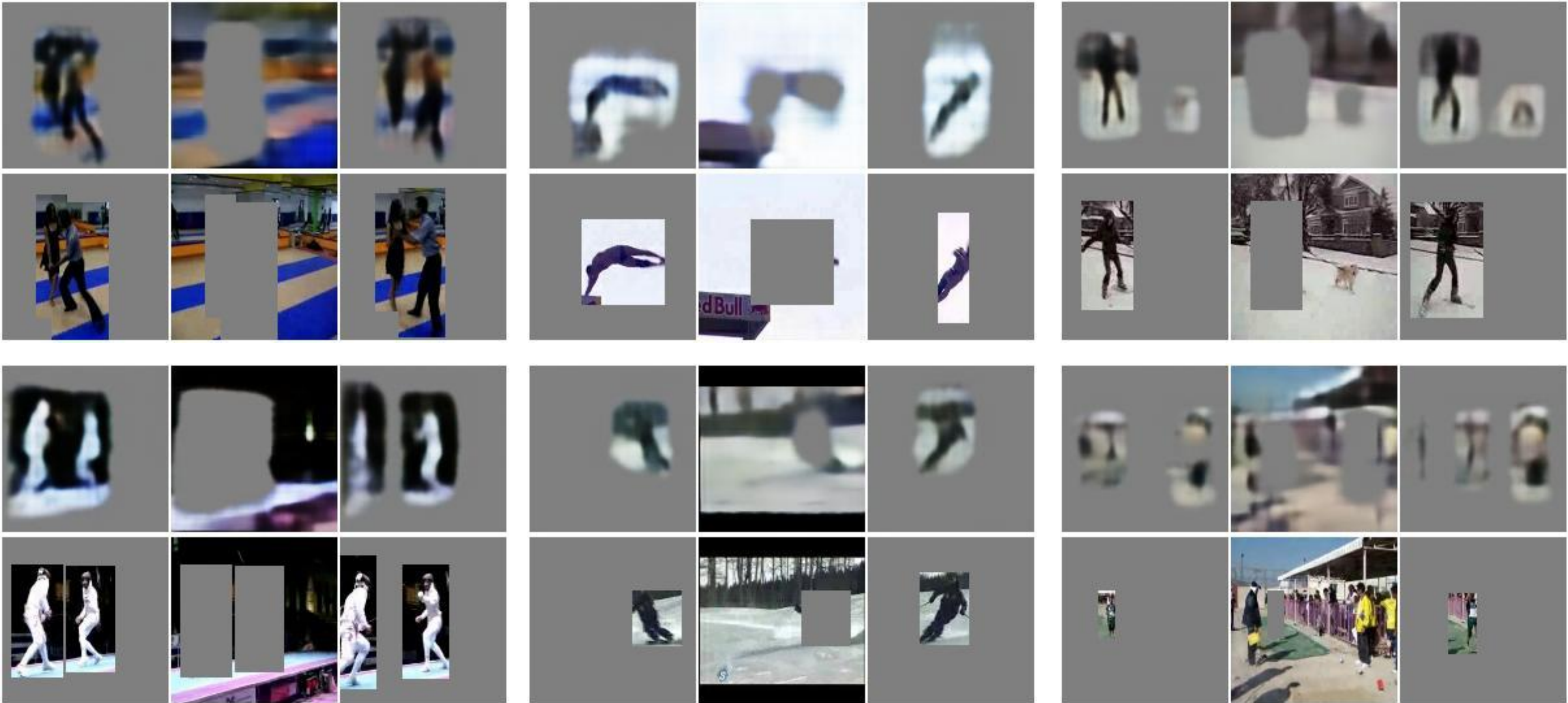}\\
\caption{Reconstruction results on the test set. For each example, the top row shows the reconstruction while the bottom one contains the ground truth. Each column shows the segmentation of foreground in first frame, background in first frame and foreground in last frame, respectively.}
\label{fig:recon_results}
\end{figure*}

We tested our model on test set for reconstruction task. For better demonstrating the efficiency of our proposed pretraining pipeline, we also trained the network to do action recognition with pretrained features.

\textbf{Reconstruction task:} reconstruction results on test set are shown in Figure \ref{fig:recon_results}. From these results, we can clearly see that the network already can predict similar foreground segmentation as ground truth. However, the image reconstructions are still blurry. We argue that this is due to the properties of the L1 loss we are adopting \cite{mathieu2015deep}. One interesting fact is that the network has learned to generalize foreground to some other moving objects in the scene even though they are not included in the annotations. For example, the result shown in the top-right corner: instead of only segmenting the person, the dog walking beside the person is also included. This fact suggests that the network has successfully learned to identify foreground from motion cues.

Besides from foreground and background features, these results also demonstrate a good extraction of motion features. The learned motion features contain both global motions, e.g.~transition of foreground, and local motions, e.g.~change of human pose. In the bottom-center result, the generated kernels from motion feature successfully shift the object from right to the middle and change its gesture.

\textbf{Action recognition:} a good pretraining pipeline should show better performance on some typical discriminative tasks than random initialization, especially when training data is scarce \cite{misra2016shuffle,luo2017motionprediction,pathak2017learning,vondrick2016generating,wang2015unsupervised}. We also conducted comparative experiments on the task of action recognition. By discarding the decoders in our framework and training a linear softmax layer on top of the disentangled features, we can obtain a simple network for action recognition. For the first experiment, we first pretrain our encoder on the subset of UCF-101 with the settings discussed above and then fine-tune the whole action recognition network with added softmax layer on the same subset. As baselines, we trained another two action recognition networks, one with all weights initialized randomly and another one pretrained with an unsupervised autoencoder architecture. This autoencoder shared the same 3D convolutional encoder architecture with ours, while its decoder was the mirrored version of the encoder but replacing the pooling operations with convolutions.

\begin{figure}
\centering
\includegraphics[width=\linewidth]{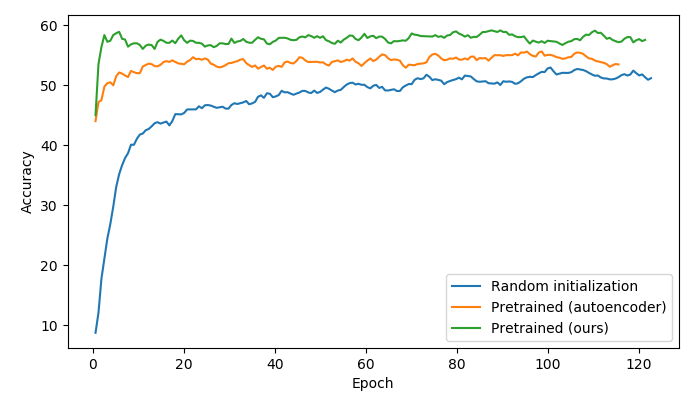}\\
\caption{Action recognition results on validation set. This figure shows the accuracy of each method on validation set during the training time.}
\label{fig:val_results}
\end{figure}

During training, we observed that our pretrained model reached 90\% accuracy on training set immediately after one epoch while the randomly initialized network took 130 epochs to achieve it. All three models reached around 96\% accuracy at the end of training and encountered severe overfitting problems. The accuracy of different methods on the validation set during training time is shown in Figure \ref{fig:val_results}. The best accuracy obtained on the test set with our pretrained model is 62.5\%, while it drops to 52.2\% and 56.8\% respectively when using a random initialization and autoencoder as pretraining scheme, as shown in Table \ref{tab:test_acc}. We observe a margin of more than 10\% on accuracy between our proposed method and random initialization on both validation set and test set. This further demonstrates that with our proposal, the network can learn features that generalize better. These results are specially promising given the small amount of data used during pretraining, which is just a fraction of UCF-101. While this demonstrates the efficiency of the approach, using a larger dataset for pretraining should provide additional gains and better generalization capabilities.

\begin{table}
  \centering
  \caption{Action recognition accuracy of different methods on the test subset of UCF-101.}
  \label{tab:test_acc}
  \resizebox{0.6\linewidth}{!}{
    \begin{tabular}{cc}
      \toprule
      \textbf{Method}  & \textbf{Accuracy}  \\ 
      \midrule
      Random initialization & 52.2\%  \\ 
      \midrule
      Pretrained (autoencoder) & 56.8\%  \\ 
      \midrule
      Pretrained (ours)  & \textbf{62.5\%}  \\
      \bottomrule
    \end{tabular}
  }
\end{table}

\section{Conclusions}

This work has proposed a novel framework towards an unsupervised learning of video features capable of disentangling of motion, foreground and background.
Our method mostly exploits motion in videos and is inspired by human perceptual grouping with motion cues. 
Our experiments using ground truth boxes render convincing results on both frame reconstruction and action recognition, showing the potential of the proposed architecture.

However, multiple aspects still need to be explored in our work. As our plans for the future work, we decide to (1) introduce unsupervised learning for foreground segmentation as well, as proposed in uNLC \cite{pathak2017learning}; (2) train with a larger amount of unlabeled data; (3) introduce adversarial loss to improve the sharpness of the reconstructed frames \cite{mathieu2015deep}; and (4) fill the gap of absent motion features between the first frame and the last frame by reconstructing any random frame in the clip.

Our model and source code are publicly available at \url{https://imatge-upc.github.io/unsupervised-2017-cvprw/} .
\section*{Acknowledgments}
The Image Processing Group at UPC is supported by the project TEC2013-43935-R and TEC2016-75976-R, funded by the Spanish Ministerio de Economia y Competitividad and the European Regional Development Fund (ERDF). The Image Processing Group at UPC is a SGR14 Consolidated Research Group recognized and sponsored by the Catalan Government (Generalitat de Catalunya) through its AGAUR office. 
The contribution from the Barcelona Supercomputing Center has been supported by project TIN2015-65316 by the Spanish Ministry of Science and Innovation contracts 2014-SGR-1051 by Generalitat de Catalunya.

{\small
\bibliographystyle{ieee}
\bibliography{egbib}

\begin{thebibliography}{10}\itemsep=-1pt

\bibitem{du2015c3d}
T.~Du, B.~Lubomir, F.~Rob, T.~Lorenzo, and P.~Manohar.
\newblock {C3D:} generic features for video analysis.
\newblock In {\em ICCV}, 2015.

\bibitem{THUMOS15}
A.~Gorban, H.~Idrees, Y.-G. Jiang, A.~Roshan~Zamir, I.~Laptev, M.~Shah, and
  R.~Sukthankar.
\newblock {THUMOS} challenge: Action recognition with a large number of
  classes.
\newblock \url{http://www.thumos.info/}, 2015.

\bibitem{lecun15nature}
Y.~LeCun, Y.~Bengio, and G.~Hinton.
\newblock Deep learning.
\newblock In {\em Nature}, 2015.

\bibitem{luo2017motionprediction}
Z.~Luo, B.~Peng, D.-A. Huang, A.~A. Alahi, and L.~Fei-Fei.
\newblock Unsupervised learning of long-term motion dynamics for videos.
\newblock In {\em CVPR}, 2017.

\bibitem{mathieu2015deep}
M.~Mathieu, C.~Couprie, and Y.~LeCun.
\newblock Deep multi-scale video prediction beyond mean square error.
\newblock In {\em ICLR}, 2016.

\bibitem{misra2016shuffle}
I.~Misra, C.~L. Zitnick, and M.~Hebert.
\newblock Shuffle and learn: unsupervised learning using temporal order
  verification.
\newblock In {\em ECCV}, 2016.

\bibitem{pathak2017learning}
D.~Pathak, R.~Girshick, P.~Doll{\'a}r, T.~Darrell, and B.~Hariharan.
\newblock Learning features by watching objects move.
\newblock In {\em CVPR}, 2017.

\bibitem{elizabeth90cognitivemotion}
E.~S.~Spelke.
\newblock Principles of object perception.
\newblock In {\em Cognitive Science}, 1990.

\bibitem{simonyan2014two}
K.~Simonyan and A.~Zisserman.
\newblock Two-stream convolutional networks for action recognition in videos.
\newblock In {\em NIPS}, 2014.

\bibitem{UCF101}
K.~Soomro, A.~Roshan~Zamir, and M.~Shah.
\newblock {UCF101}: A dataset of 101 human actions classes from videos in the
  wild.
\newblock In {\em CRCV-TR-12-01}, 2012.

\bibitem{srivastava2015unsupervised}
N.~Srivastava, E.~Mansimov, and R.~Salakhutdinov.
\newblock Unsupervised learning of video representations using lstms.
\newblock In {\em ICML}, 2015.

\bibitem{villegas2017decomposing}
R.~Villegas, J.~Yang, S.~Hong, X.~Lin, and H.~Lee.
\newblock Decomposing motion and content for natural video sequence prediction.
\newblock In {\em ICLR}, 2017.

\bibitem{vondrick2016generating}
C.~Vondrick, H.~Pirsiavash, and A.~Torralba.
\newblock Generating videos with scene dynamics.
\newblock In {\em NIPS}, 2016.

\bibitem{wang2015unsupervised}
X.~Wang and A.~Gupta.
\newblock Unsupervised learning of visual representations using videos.
\newblock In {\em ICCV}, 2015.

\bibitem{visualdynamics16}
T.~Xue, J.~Wu, K.~L. Bouman, and W.~T. Freeman.
\newblock Visual dynamics: Probabilistic future frame synthesis via cross
  convolutional networks.
\newblock In {\em NIPS}, 2016.

\end{thebibliography}
}

\end{document}